\documentclass[11pt]{article}
\usepackage{coling2020}
\usepackage{times}
\usepackage{url}
\usepackage{latexsym}
\usepackage{times}

\usepackage{CJKutf8}
\usepackage{amsmath}
\usepackage{booktabs}
\usepackage{times}  
\usepackage{helvet} 
\usepackage{courier}
\usepackage{graphicx}

\colingfinalcopy 

\title{Towards Fast and Accurate Neural Chinese Word Segmentation with Multi-Criteria Learning}

\author{Weipeng Huang\thanks{~~Equal contribution.} \quad Xingyi Cheng$^{*}$\thanks{~~Corresponding author.}  \quad Kunlong Chen  \quad Taifeng Wang \quad Wei Chu   \\
Ant Group \\
{\tt  \{weipeng.hwp,fanyin.cxy, }\\
{\tt  kunlong.ckl,taifeng.wang,weichu.cw\}@antgroup.com} \\
}

\date{}

\begin{document}
\maketitle
\begin{abstract}
		The ambiguous annotation criteria lead to divergence of Chinese Word Segmentation (CWS) datasets in various granularities. 
		Multi-criteria Chinese word segmentation aims to capture various annotation criteria among datasets and leverage their common underlying knowledge. 
		In this paper, we propose a domain adaptive segmenter to exploit diverse criteria of various datasets. Our model is based on Bidirectional Encoder Representations from Transformers (BERT), which is responsible for introducing open-domain knowledge. 
		Private and shared projection layers are proposed to capture domain-specific knowledge and common knowledge, respectively. 
		We also optimize computational efficiency via distillation, quantization, and compiler optimization. Experiments show that our segmenter outperforms the previous state of the art (SOTA) models on 10 CWS datasets with superior efficiency.
\end{abstract}

	\section{Introduction}
	
	Chinese Word Segmentation (CWS) is typically regarded as a low-level NLP task.
	Unlike English and French that uses the space token to separate the words, Chinese is a kind of polysynthetic languages where compounds are developed from indigenous morphemes~\cite{jernudd2011politics,gong2017multi}. 
	The ambiguous distinction between morphemes and compound words leads to the cognitive divergence of word concepts. 
	Consequently, the labeled datasets seriously diverge due to the annotation inconsistency, resulting in multi-grained compounds. 
	As shown in Table~\ref{diverse-criteria}, given a sentence ``\begin{CJK*}{UTF8}{gbsn}刘国梁赢得世界冠军\end{CJK*}'' (Liu Guoliang wins the world championship), the two commonly used corpora, i.e., PKU’s People’s Daily (PKU) and Penn Chinese Treebank (CTB), use different segmentation criteria.
	
	In practice, a segmenter usually provides multiple configures with different granularities to better serve various downstream tasks. 
	Fine-grained criterion is able to reduce the vocabulary, thereby relieves the sparseness issue. 
	On the other hand, coarse-grained words provide more specific meanings, which may benefit the domain-specific tasks. 

	In recent years, several multi-criteria learning methods for CWS have been proposed to explore the common knowledge of heterogeneous datasets. 
	By utilizing the information across all corpora, multi-criteria learning methods can boost the out-of-vocabulary (OOV) recalls as well as practical performance~\cite{qiu2013joint,chao2015exploiting,chen2017adversarial}. 
	Despite its effectiveness, there still are three unresolved issues. 
	(1) Even with multiple datasets, the data is still limited to provide adequate linguistic knowledge. 
	(2) Learning from a dataset is likely to hurt the others as the segmentation follows inconsistent criteria. 
	(3) The advanced models, e.g., Bi-LSTM-CRF~\cite{ma2018state}, are computationally expensive.
	They are based on the recurrent neural networks (RNNs). 
	Since RNNs are auto-regressive and the computation cannot be completed in parallel, their applications are usually limited due to the poor computational efficiency.
	In fact, the inference speed is heavily required for the CWS system as it serves as a fundamental module of NLP pipelines. 
	For example, search engines generally can only afford to spend tens of milliseconds or even milliseconds in CWS. 

	
	To alleviate the limitations of existing methods, we propose a multi-criteria method for CWS. 
	Recent studies~\cite{yang2017neural,ma2018state,wang2019unsupervised} pointed out that exploiting external knowledge can improve the CWS accuracy.
	Based on this observation, we adopt BERT~\cite{vaswani2017attention,devlin2018bert} as the backbone to extract the open-domain knowledge.
	On the top of BERT, private projection layers and shared projection layers are used to capture domain-specific knowledge and common underlying knowledge respectively. 
	
	To make it more practicable, three techniques, i.e., knowledge distillation, numeric quantization and compiler optimization, are adopted to accelerate our segmenter.
	The BERTology analysis~\cite{clark2019does,jawahar2019does,xu2020symmetric} indicated that the representations from different layers of BERT capture specific meanings.
	It is sufficient to use the representations from a middle layer for the CWS task~(see section~\ref{sec:layer_analysis} for detailed analysis).
	To make the best use of BERT, the knowledge distillation method proposed by~\cite{hinton2015distilling} is utilized.
	
	Simultaneously, we also adopt quantization and compiler optimization techniques to improve the scalability. 
	Experiments show that our method not only significantly outperforms the best known results on 10 CWS datasets with better efficiency.

	The contributions could be summarized as follows.
	
	\begin{itemize}
		\item BERT with a domain projection layer on the top is employed to capture heterogeneous segmentation criteria and common underlying knowledge. To our knowledge, it is the first time to utilize pre-trained model in CWS.
		
		\item{We visualize the BERT layers and attention scores to give an insight into linguistic information within CWS.}
		
		\item Model acceleration techniques including distillation, quantization and compiler optimization, are adopted to improve the segmentation speed.
		
		\item Experimental results show that our model outperforms previous results on 10 CWS corpora with different segmentation criteria.

	\end{itemize}

	\begin{table*}
		\centering
		\resizebox{.55\textwidth}{!}{
		\begin{tabular}{c|c|c|c|c|c}
			\toprule Criteria & Liu & Guoliang & wins & \multicolumn{2}{|c}{the world championship}\\
			\hline CTB & \multicolumn{2}{|c|}{\begin{CJK*}{UTF8}{gbsn}刘国梁\end{CJK*}} & \begin{CJK*}{UTF8}{gbsn}赢得 \end{CJK*} &  \multicolumn{2}{|c}{\begin{CJK*}{UTF8}{gbsn}世界冠军\end{CJK*}}\\
			\hline PKU & \begin{CJK*}{UTF8}{gbsn}刘\end{CJK*} & \begin{CJK*}{UTF8}{gbsn}国梁\end{CJK*} & \begin{CJK*}{UTF8}{gbsn}赢得\end{CJK*} & \begin{CJK*}{UTF8}{gbsn}世界\end{CJK*} & \begin{CJK*}{UTF8}{gbsn}冠军\end{CJK*}\\
			\toprule
		\end{tabular}
		}
		\vspace*{-1mm}
		\caption{Diverse segmentation criteria.}
		\label{diverse-criteria}
	\end{table*}

	\section{Model Description} 
	
	Current neural CWS models usually consist of three components: a character embedding layer, a feature extraction layer and a CRF tag inference layer. 
	To equip our model with the ability of multi-criteria learning, we insert an extra domain projection layer before the inference layer, as shown in Figure~\ref{model-framwork}.  
	In this section, we describe the proposed model architecture and the objective function in detail.
	
	\subsection{Feature Extraction Layer}
\begin{figure*}[!t]
		\centering
		\includegraphics[width=0.5\textwidth]{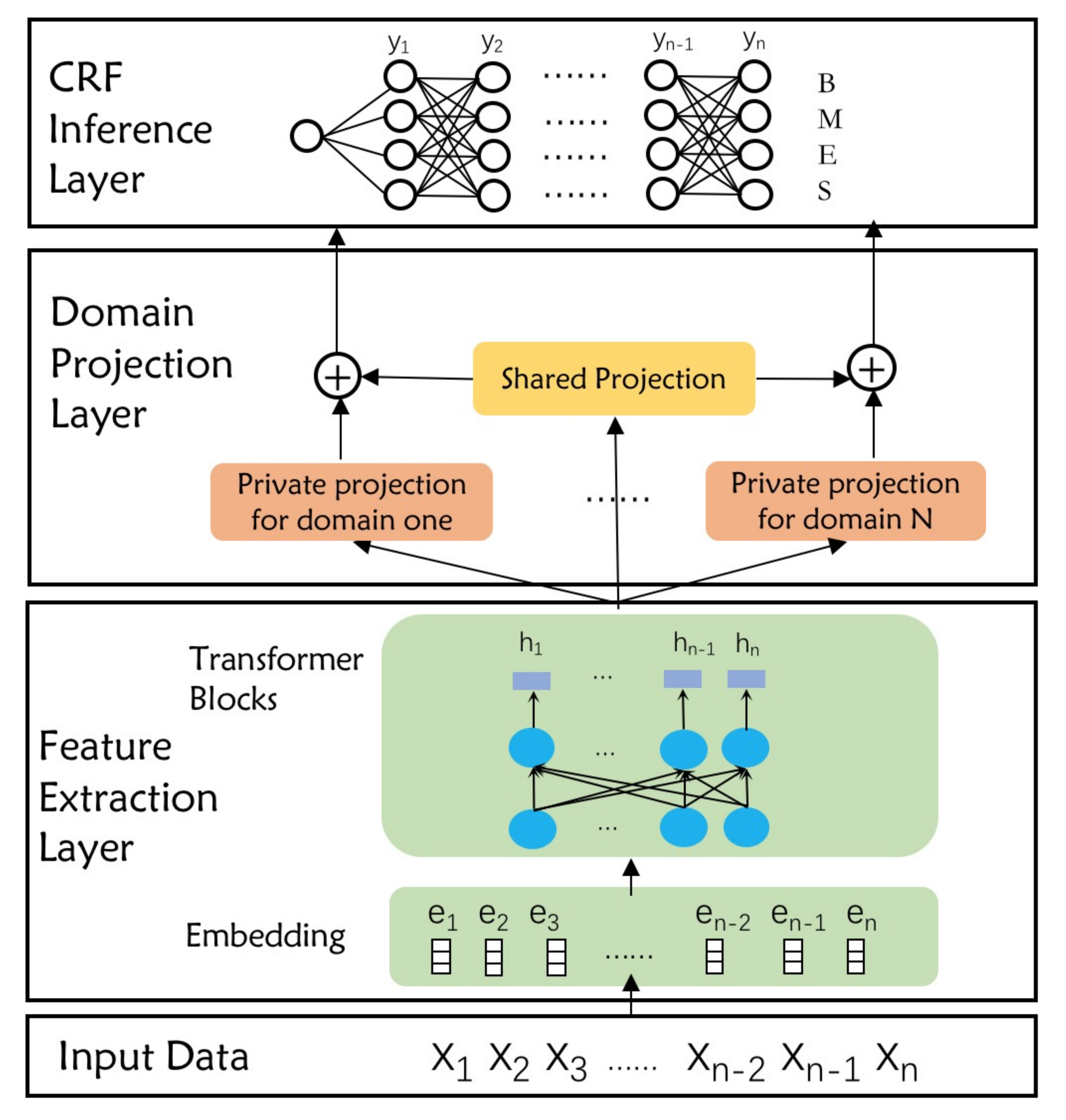}
		\vspace*{-6mm}
		\caption{The architecture of the proposed model, stacked with a feature extraction layer, a domain projection layer and a CRF tag inference layer.}
		\label{model-framwork}
	\end{figure*}

	We employ BERT ~\cite{vaswani2017attention,devlin2018bert} to extract feature for the input sequence. 
	BERT is of critical importance for the word segmentation task. 
	As shown in Figure~\ref{model-framwork}, the characters are first mapped into embedding vectors and then fed into several transformer blocks. Compared with Bi-LSTM which processes the sequence step by step, the transformer learns features in parallel for all time-steps so that the decoding speed can be accelerated. 
	However, the original BERT with 12 transformer layers is still too heavy to be applied in the real-world word segmentation application. 
	To speed up both the fine-tuning and inference procedures, we make further optimization as discussed in Section~\ref{sec:section-speed-optimization}. 
	
	Given a sentence $X ={\left \{ x_{1},x_{2},...,x_{n} \right \}}$,  we first map each character  $x_{i}$ into embedding vector $\textbf{e}_{i}$. The embedding vectors are then fed into BERT to get the feature representations:
	\begin{equation}
		\textbf{h}_{i}=\texttt{BERT}(\textbf{e}_{1},\textbf{e}_{2},...,\textbf{e}_{n};\theta)\,, 
	\end{equation}
	where $\theta$ denotes all the parameters in BERT model.
	
	\subsection{Domain Projection Layer}
	As shown in Table~\ref{diverse-criteria}, the same sentence can be segmented into different words according to different dataset criteria. If we simply combine the datasets to train a single model, the model will be confused by diverse criteria and thus hurt the performance. 
	Traditional methods train an individual model for each dataset, which results in huge deployment costs in practice. 
	
	Inspired by previous works \cite{chen2017adversarial,peng2017}, we propose a domain projection layer to enable our model to adapt the datasets with various criteria. 
	The domain projection layer helps to capture heterogeneous segmentation criteria of each dataset. 
	There are many variations of the projection layer; in this paper, we use a linear transformation layer, which is simple but effective for this task. 
	As shown in Figure~\ref{model-framwork}, we introduce an extra shared projection layer to learn common knowledge from datasets.
	The computation graph excludes domain-specific projections and reserves the shared projection to segment with standard criteria. 
	
	Formally, with the domain projection layer, we can obtain the domain-specific representation and shared representation:
	\begin{align}
		\textbf{h}_{domain} &= \textbf{W}_{domain}^{T}\textbf{h} + \textbf{b}_{domain}, \\
		\textbf{h}_{shared} &= \textbf{W}_{shared}^{T}\textbf{h} + \textbf{b}_{shared},
	\end{align}
	where $ \textbf{W}_{domain}^{T} \in \mathbf{R}^{d_{h}\times d_{h}}$, $ \textbf{W}_{shared}^{T} \in \mathbf{R}^{d_{h}\times d_{h}}$ ,  $\textbf{b}_{domain} \in \mathbf{R}^{d_{h}}$  and $\textbf{b}_{shared} \in \mathbf{R}^{d_{h}}$ are  trainable parameters. $d_{h}$ is the dimension of $\textbf{h}$.
	
	\subsection{Inference Layer}
	We treat the CWS task as a character-based sequence labeling problem. Each character in a sentence $X ={\left \{ x_{1},x_{2},...,x_{n} \right \}}$  is labelled as one of $\mathcal{L} = {\left \{ B,M,E,S \right \}}$,  indicating the begin, middle, end of a word, and a word with single character. As shown in Figure~\ref{model-framwork}, the output of domain-specific projection and shared projection are concatenated, then fed into a first-order linear-chain conditional random fields (CRF) layer \cite{Lafferty2001} to inference these tags.
	
	Formally, the probability of a label sequence formalized as:
	\begin{equation}
		p(Y|X)=\frac{\Psi(Y|X)}{\sum_{Y^{'}\in \mathcal{L}^{n}}\Psi(Y^{'}|X)}\,,
	\end{equation}
	where $\Psi(Y|X)$ is potential function:
	\begin{align}
		\Psi(Y|X)&=\prod_{i=2}^{n}\psi(X,i,y_{i-1},y_{i}) \,, \\
		\psi(X,i,y_{i-1},y_{i})&=\exp(s(X,i)_{y_{i}}+\textbf{b}_{y_{i-1}y_{i}})\,,
	\end{align}
	where $y\in {\left \{ B,M,E,S \right \}}$ is the tag label, $\textbf{b} \in \mathbf{R}^{|\mathcal{L}| \times |\mathcal{L}| }$ is trainable parameter and $ \textbf{b}_{y_{i-1}y_{i}} $ means transition from tag $y_{i-1}$ to $y_{i}$.  Score function $ s(X,i) $ is output of the projection layer at $ i_{th}$ character, which assigns score for each label on tagging the  $ i_{th}$  character:
	\begin{equation}
		s(X,i) = \textbf{W}_{s}^{T}[\textbf{h}_{domain};\textbf{h}_{shared}] + \textbf{b}_{s} ,
	\end{equation}
	where $[\textbf{h}_{domain};\textbf{h}_{shared}]$ is the concatenation of domain-specific projection and shared projection, $\textbf{W}_{s}^{T} \in \mathbf{R}^{2d_{h}\times|\mathcal{L}|}$ and $\textbf{b}_{s} \in \mathbf{R}^{|\mathcal{L}|}$ are  trainable parameters.

	The inference can be achieved by maximizing the posterior probability:
	\begin{equation}
		\label{eq-2}
		Y^{*} = \arg\max p(Y|X)\,.
	\end{equation}

	\subsection{Objective Function}
	The parameters of the network are trained to maximize the conditional log-likelihood of true labels on the dataset. The objective function $\mathcal{J}_{seg}$ is computed as :
	\begin{equation}
		\mathcal{J}_{seg}(\Theta) = \sum_{j} \log p(Y^{(j)}|X^{(j)};\Theta)\,,
	\end{equation}
	where $\Theta$  denotes all the parameters in the model, $(X^{(j)}, Y^{(j)})$ denotes the $j_{th}$ sample in the datasets.
	The total loss for multi-criteria learning is the combination of loss in each datasets.

	\section{Model Acceleration}
	\label{sec:section-speed-optimization}
	
	Neural CWS models improve the performance by increasing the model complexity, which however harms the decoding speed and limits their real-world application. 
	To bridge this gap, we apply model acceleration techniques as follows.
	
	\subsection{Distillation} 
	\label{sec:distill}
	
	To balance computational cost and segmentation accuracy, we distill knowledge~\cite{ba2014deep,hinton2015distilling} from BERT into a smaller transformer network.
	And the supervised fine-tuning on the datasets are performed jointly. 
	Recent analysis~\cite{clark2019does,jawahar2019does} show that the layers of BERT provide phrase-level information, the middle layers extract syntactic features and the top layers are capable of handling semantic features. CWS is essentially a syntactic chunking task and heavily relies on lexical and syntactic features. Therefore, we turn to use bottom-to-middle layers as the backbone to learn annotations and jointly distill the top layer of BERT. Specifically, the original Chinese BERT with 12 layers serve as a teacher, and a truncated  (3 or 6 layers)  BERT learns from the teacher as a student. 
	The teacher network and student network differ in the feature extraction layer of our proposed model shown in Figure~\ref{model-framwork}. 
	
	To distill the original BERT, we add a logits-regression objective by minimizing the square loss between the normalized logits from the teacher model and the logits from the student model. The distillation loss is formulated as:
	\begin{equation}
		\small
		\mathcal{J}_{dis}(\Theta_{s},\Theta_{t}) = \frac{1}{2T} \sum_{j=0}^{M} \sum_{i=0}^{N} \Big( \frac{\textbf{h}_{s}^{(j,i)}}{||\textbf{h}_{s}^{(j,i)}||_{2}}- \frac{\textbf{h}_{t}^{(j,i)}}{||\textbf{h}_{t}^{(j,i)}||_{2}}\Big) ^2,
	\end{equation}
	where $\Theta_{s}$, $\Theta_{t}$  denote parameters in the student network and teacher network, $M$ denotes the number of samples in the datasets, $N$ denotes the sequence length, $\textbf{h}_{s}$
	,$\textbf{h}_{t}$ denote the logits extracted from student network and teacher network respectively. During the distillation process, the parameters $\Theta_{t}$ are frozen. 
	
	Combining the segmentation loss and the distillation loss, the overall loss is:
	\begin{equation}
		\mathcal{J}(\Theta_{s},\Theta_{t}) = \mathcal{J}_{seg}(\Theta_{s}) + \alpha \mathcal{J}_{dis}(\Theta_{s},\Theta_{t})  ,
	\end{equation}
	where $\alpha$ is a hyper-parameter to trade off these two loss function.
	
	\subsection{Quantization} 
	Quantization methods also have been investigated for network acceleration. 
	These approaches are mainly  categorized into two groups: scalar and vector quantization \cite{gong2014compressing}, fixed-point quantization \cite{gupta2015deep}.  Traditional neural networks implementation use 32-bit single-precision floating-point for both weights and activation, resulting in a cost of the substantial
	increase in computation and model storage resources. Therefore, we conduct fixed-point quantization to leverage NVIDIA's Volta architectural features. Fixed-point quantization was proposed to alleviate these complexities. 
	
	Specifically, in our model, half-precision (FP16) is applied on kernels of multi-head attention layers and feedforward layers, while rest parameters like embedding and normalization parameters use full precision (FP32). Gradients in fine-tuning procedures also use full precision. The quantization method not only accelerates the computation but also reduce the model size.

	\begin{table*}[t]
		\centering
		\small
		\begin{tabular}{c|ccc|c}
			\toprule
			Datasets & Training set & Development set & Testing set & Average word length \\
			\hline
			CNC & 5920K & 657K & 727K & 1.52 \\
			\hline
			AS & 4903K & 546K & 122K & 1.51 \\
			\hline
			MSR & 2132K & 235K & 106K & 1.68 \\
			\hline
			CITYU & 1309K & 146K & 40K & 1.62 \\
			\hline
			PKU & 994K & 115K & 104K & 1.61 \\
			\hline
			CTB6 & 641K & 59K & 81K & 1.63 \\
			\hline
			SXU & 476K & 53K & 113K & 1.57 \\
			\hline
			UD & 100K & 11K & 12K & 1.49 \\
			\hline
			ZX & 79K & 9K & 34K & 1.42 \\
			\hline
			WTB & 15K & 1K & 2K & 1.53 \\
			\bottomrule
			
		\end{tabular}
		\vspace*{-1mm}
		\caption{Details of the ten datasets: the number of words in the training set, development set and testing set, the average word length (char/word) of each dataset.
		}
		\label{dataset-detail} 
	\end{table*}

	\begin{table*}[t!]
		\centering
		\small
		\begin{tabular}{c|cccccccccc}
			\toprule
			& PKU & MSR & AS & CITYU & CTB6 & SXU & UD & CNC & WTB & ZX\\
			\hline
			
			\cite{yang2017neural} & 96.3 & 97.5 & 95.7 & 96.9 & 96.2 & - & - & - & - & -  \\
			\cite{chen2017adversarial} & 94.3 & 96.0 & 94.6 & 95.6 & 96.2 & 96.0 & - & - & - & -  \\
			\cite{Xu:2016} & 96.1 & 96.3 & - & - & 95.8 & - & - & - & - & -  \\
			
			\cite{ma2018state} & 96.1 & 98.1 & 96.2 &  97.2 & 96.7 & - &  96.9 & - & - & -  \\
			\cite{gong2018switch} & 96.2 & 97.8 & 95.2 & 96.2 & 97.3 &  97.2 & - & - & - & -  \\
			\cite{zhou2019multiple} & 96.2 & 97.0 & 96.9 & 97.1 & 95.2 & - & - & - & - & -  \\
			\cite{He:2019} & 96.0 & 97.2 & 95.4 & 96.1 &  96.7 &  96.4 & 94.4 &  97.0 &  90.4 &  95.7  \\
			\hline
			

			Ours (Student-3 layer)   &  96.7 &   97.9 &   96.8 &   97.6 &   97.5 &   97.3 &   97.4 &  97.1 &   93.1 &  97.0  \\
			Ours (Student-3 layer+FP16)   & 96.6 &  98.0 & 96.6 & 97.5 & 97.4 &  97.3 &  97.3 & 97.1 & 92.7 &  96.8  \\
			Ours (Student-6 layer)   &  97.2 &   98.3 &   97.0 &   97.7 &  97.7 &  \bf 97.5 &  \bf 97.8 &  97.2 &   93.0 & \bf 97.1\\
			Ours (Student-6 layer+FP16)  & 97.0 & 98.2 & 96.8 & \bf 97.8 &  97.7 &  97.4 & 97.7 & 97.1 &  93.1 & 96.8  \\
			Ours (Teacher-12 layer)   & \bf 97.3 &  \bf 98.5 &  \bf 97.0 &  \bf  97.8 &  \bf 97.8 &  \bf 97.5 &  \bf 97.8 & \bf 97.3 &  \bf 93.2 & \bf 97.1\\
			Ours (Teacher-12 layer+FP16)  & 97.2 & 98.3 & 96.9 & \bf 97.8 &  97.7 &   97.3 & 97.7 & 97.2 &  93.1 & 96.9  \\
			
			\bottomrule

		\end{tabular}
		\vspace*{-1mm}
		\caption{\label{f1-results}Comparison among the state-of-the-art performance on different datasets (F1-score, \%).
		}
	\end{table*}
	
	\subsection{Compiler Optimization} 
	XLA (Accelerated Linear Algebra) is a domain-specific compiler for linear algebra that accelerates TensorFlow models by optimizing one's computations. It provides an alternative mode of running TensorFlow models: it compiles the TensorFlow graph into a sequence of computation kernels generated specifically for the given model. Because these kernels are unique to the model, they can exploit model-specific information for optimization. For example, operations like addition, multiplication and reduction can be fused into a single GPU kernel.
	
	By introducing XLA into our model, graphs are compiled into machine instructions, and low-level ops are fused to improve the execution speed. For example, \texttt{batch matmul} is always followed by a transpose operation in the transformer computation graph. By fusing these two operations, the intermediate product does not need to write back to memory, thus reducing the redundant memory access time and kernel launch overhead.

	\begin{table*}
		\centering
		\begin{tabular}{c|ccc}
			\toprule
			& Precision & Recall & F1-Score  \\
			\hline
			Teacher-BERT (12 layer)  & 97.2 & 97.0 & 97.1 \\
			Student-Transformer (6 layer)  & 97.1 & 97.0 & 97.0 \\
			Student-Transformer (3 layer))  & 96.8 & 96.9 & 96.8 \\
			Student-Transformer (1 layer)  & 95.9 & 96.1 & 96.0\\
			\bottomrule
			
		\end{tabular}
		\vspace*{-1mm}
		\caption{\label{avg-prf} Average Precision, Recall, F1-score on 10 datasets for the teacher network and the student network.
		}
	\end{table*}

	\begin{table*}
		\centering
		\small
		\begin{tabular}{c|c|c|c|c|c|c|c|c|c|c|c}
			\toprule
			& AS & CITYU & CNC & CTB6 & MSR & PKU & SXU & UD & WTB & ZX & All Datasets \\
			\hline
			AS & 0.0 & 13.9 & 14.8 & 8.4 & 7.1 & 6.4 & 4.5 & 2.3 & 0.4 & 0.9 & 30.3\\
			\hline
			CITYU & 33.5 & 0.0 & 31.4 & 14.0 & 20.9 & 17.5 & 9.8 & 4.0 & 0.8 & 2.4 & 50.5 \\
			\hline
			CNC & 14.8 & 6.2 & 0.0 & 4.2 & 7.7 & 6.0 & 2.7 & 1.7 & 0.2 & 0.6 & 25.1 \\
			\hline
			CTB6 & 47.3 & 31.2 & 40.5 & 0.0 & 28.0 & 24.6 & 15.4 & 7.1 & 1.2 & 3.1 & 63.9 \\
			\hline
			MSR & 18.4 & 10.5 & 27.3 & 8.0 & 0.0 & 10.7 & 5.9 & 2.3 & 0.1 & 1.3 & 36.0 \\
			\hline
			PKU & 39.6 & 31.7 & 50.1 & 25.3 & 35.2 & 0.0 & 15.7 & 8.6 & 0.7 & 3.9 & 67.3 \\
			\hline
			SXU & 49.8 & 39.7 & 56.2 & 29.3 & 41.8 & 36.6 & 0.0 & 10.9 & 1.7 & 4.9 & 73.2 \\
			\hline
			UD & 55.3 & 43.2 & 57.5 & 37.6 & 45.1 & 37.2 & 30.2 & 0.0 & 3.8 & 6.2 & 70.2 \\
			\hline
			WTB & 76.5 & 71.0 & 79.2 & 64.3 & 72.5 & 69.8 & 65.1 & 41.6 & 0.0 & 22.0 & 85.1 \\
			\hline
			ZX & 71.4 & 49.8 & 71.8 & 43.0 & 53.1 & 44.2 & 35.3 & 20.1 & 6.9 & 0.0 & 81.1 \\
			\bottomrule
			
		\end{tabular}
		\vspace*{-1mm}
		\caption{Each row indicates the rate of OOV words ( \% ) in a dataset appear in other datasets. 
		}
		\label{oov-in-other}
	\end{table*}

	\begin{table*}
		\centering
		\small
		\begin{tabular}{c|c|cccccccccc|c}
			\toprule
			& & PKU & MSR & AS & CITYU & CTB6 & SXU & UD & CNC & WTB & ZX & Avg\\
			\hline
			F1  & single-criteria  & 94.7 & 95.3 &  95.2 & 95.7 & 95.4 & 94.4 & 94.6 &  96.3 & 89.9 & 94.4  & 94.5\\
			Score & multi-criteria  & \bf 96.7 & \bf 97.9 & \bf 96.7 & \bf 97.6 & \bf 97.5 & \bf 97.3 & \bf 97.4 &  \bf 97.1 &  \bf 93.1 & \bf 97.0 & \bf 96.8  \\
			\hline
			OOV  & single-criteria  & 74.8 & 78.0 & \bf 78.3 & 83.7 & 62.8 & 80.1 & 73.6 &  64.2 & 73.9 & 74.8 & 74.2 \\
			Recall  & multi-criteria  & \bf 81.6 & \bf 84.0 & 77.3 & \bf 90.1 & \bf 89.4 & \bf 85.7 & \bf 91.6 & \bf 65.0 & \bf 82.9 & \bf 89.1 & \bf 83.6\\
			\bottomrule
			
		\end{tabular}
		\vspace*{-1mm}
		\caption{\label{oov-results}OOV recall(\%), F1 Score (\%) achieved with multi-criteria learning and single-criteria learning. The number of transformer layer is set to 3 for both single-criteria and multi-criteria learning.
		}
	\end{table*}

	\section{Experiments}
	
	\subsection{Experimental Settings}
	All experiments are implemented on the hardware with Intel(R) Xeon(R) CPU E5-2682 v4 @ 2.50GHz and NVIDIA Tesla V100.
	
	\textbf{Datasets.} We evaluate our model on ten standard Chinese word segmentation datasets: MSR,PKU,AS,CITYU from SIGHAN 2005 bake-off task \cite{Emerson:2005}. SXU from SIGHAN 2008 bake-off task \cite{MOE:2008}. Chinese Penn Treebank 6.0 (CTB6) from \cite{Xue:2005}. Chinese Universal Treebank (UD) from the Conll2017 shared task \cite{Zeman:2017}. WTB \cite{Wang:2014}, ZX \cite{Zhang:2014} and CNC corpus.  For each of the SIGHAN 2005 and 2008 dataset, we randomly select 10\% training data as the development set. For other datasets, we use official data split.  Table~\ref{dataset-detail} shows the details of the ten datasets. We can notice that the average word length (char per word) of these datasets range from 1.42 to 1.68, which reflects the diverse segmentation granularities and data distribution of these datasets. 
	
	\textbf{Preprocessing.} AS and CITYU are mapped from traditional  Chinese to simplified Chinese before segmentation. Continuous English characters and digits in the datasets are respectively replaced with a unique token. Full-width tokens are converted to half-width to handle the mismatch between training and testing set.
	
	\textbf{Hyperparameters.} The number of domain projection layer is 1, the max sequence length is set to 128. During fine-tuning, we use Adam with the learning rate of 2e-5, L2 weight decay of 0.01, dropout probability of 0.1.  For the trade-off hyperparameter $\alpha$, we had tried several value and empirically fixed it to 0.15 in the following experiment. Parameters in the feature extraction layer of teacher network and student network are initialized with pre-trained BERT\footnote{https://github.com/google-research/bert}, and all other parameters are initialized with Xavier uniform initializer.

	\textbf{Evaluation Metrics.} The goal of Chinese word segmentation is to precisely cut the input sentence into separate words. Therefore, to reach a balance of the precision ($P=\frac{\# word_{gold \cap sys}}{\# word_{sys}}$) and recall  ($R=\frac{\# word_{gold \cap sys}}{\# word_{gold}}$), we use the F1 score ($R=\frac{2PR}{P+R}$).

	\subsection{Main Results}
	\label{section-main-results}
	We distill the teacher network with the truncated BERT that compared with using the original 12 layers BERT. The average F1-score on 10 datasets using 3 layers drops slightly from 97.1\% to 96.8\% as shown in Table~\ref{avg-prf}. We suggest a student network with 3 transformer layers is a good choice to balance computational cost and segmentation accuracy.
	
	Performance of our model and recent neural CWS models are shown in Table~\ref{f1-results}. Our model outperform prior works on 10 datasets, with 13.5\%, 10.5\%, 15.8\%, 17.9\%, 14.8\%, 10.7\%, 32.2\%, 10.0\%, 29.1\%, 32.5\% error reductions on PKU, MSR, AS, CITYU, CTB6, SXU, UD, CNC, WTB, ZX datasets respectively. 
	By further applying half-precision (FP16), the accuracy reduction is minor and the model still outperforms previous SOTA results on 10 datasets.  The F1 score did not change when applying compiler optimization since it had no effect on the result of the predictions.

	\subsection{Effect of Domain Projection Layer}
	Previous work \cite{Huang:2007,ma2018state} pointed out that OOV is a major error and exploring further sources of knowledge is essential to solving this problem.  From a certain point of view, datasets are complementary to each other since OOV in a dataset may appear in other datasets. We make some analysis and the statistics are shown in Table~\ref{oov-in-other}.  Take the dataset AS for example, 13.9\% of the OOV words appear in dataset CITYU, and 30.3\% of the OOV words appear in all other datasets. 
	
	To utilize knowledge from each other to improve the OOV recall, our model performs multi-criteria learning with the domain projection layer. To evaluate this, we train the proposed model respectively on each dataset, i.e., single-criteria learning. In single-criteria learning setting, the shared projection layer is excluded and only the private projection layer is preserved for each dataset. The number of student transformer layers is set to 3 for both single-criteria and multi-criteria learning. Table~\ref{oov-results} shows that comparing with single-criteria learning, multi-criteria learning significantly improves the F1 score on all datasets, with 2.3\%  improvement on average. It also improves the OOV recall on 9/10 datasets, with 9.4\% improvement on average.

	\begin{figure*}
		\centering
		\small
		\includegraphics[width=0.8\textwidth,height=0.53\textwidth]{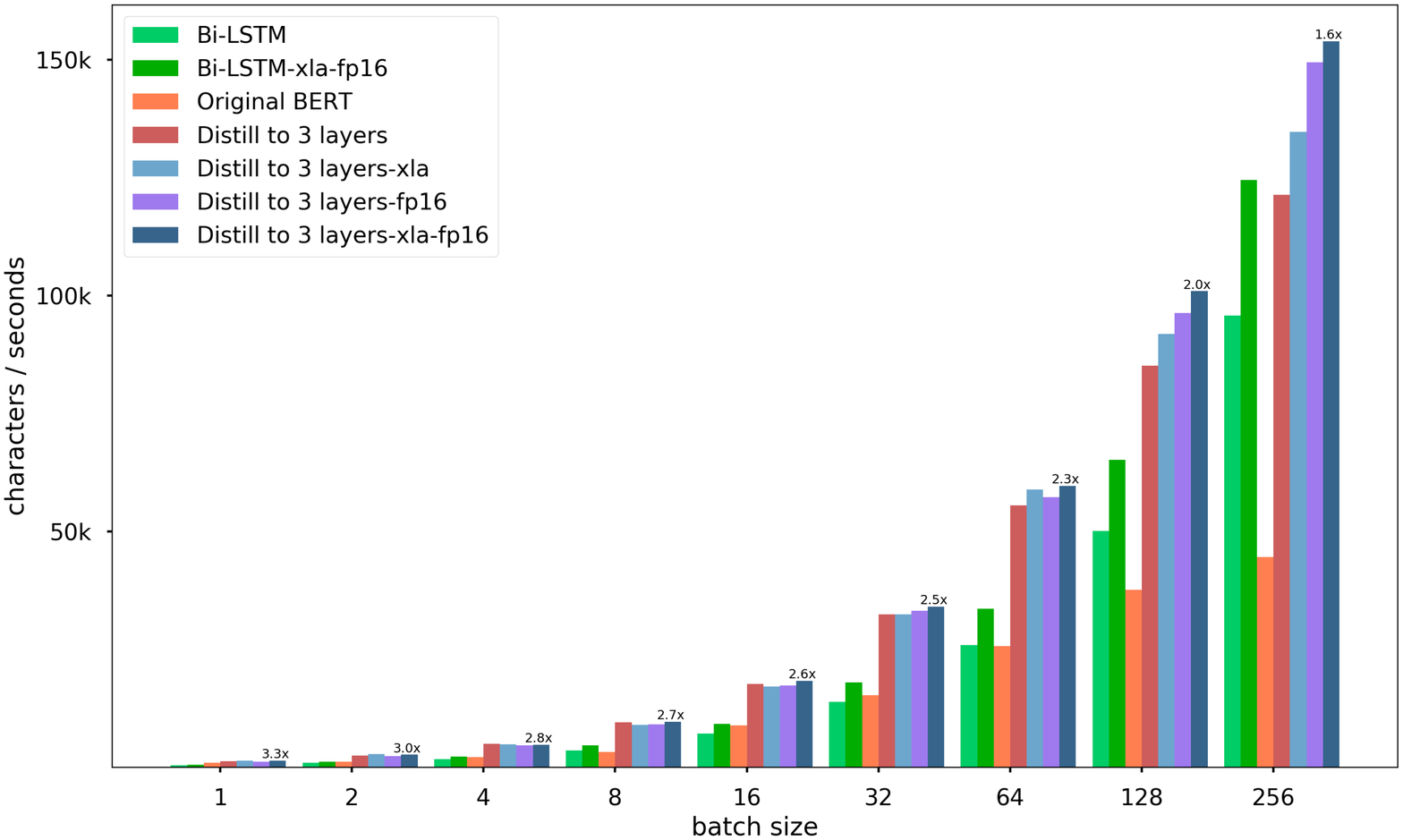}
		\vspace*{-11.5mm}
		\caption{Decoding speed w.r.t batch size. The sequence length is 64.}
		\label{decode-speed}
	\end{figure*}
	\subsection{Scalability}
	Decoding speed is essential in practice since word segmentation is fundamental for many downstream NLP tasks. Previous neural CWS models \cite{ma2018state,chen2017adversarial,gong2018switch,zhou2019multiple}  use Bi-LSTM with concatenated embedding size of 100,100,128,100 respectively. However, they did not report the decoding speed. To make a fair comparison, we set the Bi-LSTM embedding size and hidden size to 100, one hidden layer with CRF on the top. 
	
	Figure~\ref{decode-speed} shows the decoding speed with regards to batch size. Our model employed original Chinese BERT with 12 transformer layers is slower than Bi-LSTM. However, the speed can be increased by optimizations, including distillation, weights quantization, and compiler optimization. Combining all of these three techniques, our models outperform Bi-LSTM with 1.6$\times$ - 3.3$\times$ acceleration with different batch size. We try to adapt Bi-LSTM with the same computational optimization as in BERT. The result shows that Bi-LSTM achieved about 30\% acceleration by optimization. And our method still outperforms the optimized Bi-LSTM, i.e. Bi-LSTM-xla-fp16,  with 1.3$\times$ - 2.5$\times$ acceleration. Furthermore, our model is more scalable compared with the Bi-LSTM that are limited in their capability to process tasks involving very long sequences. By observing the sequence length distribution, we can search an appropriate layer number to balance F1-score and decoding speed.

	\begin{figure*}[!t]
		\centering
		\small
		\begin{tabular}{cc}
			\includegraphics[width=0.4\textwidth]{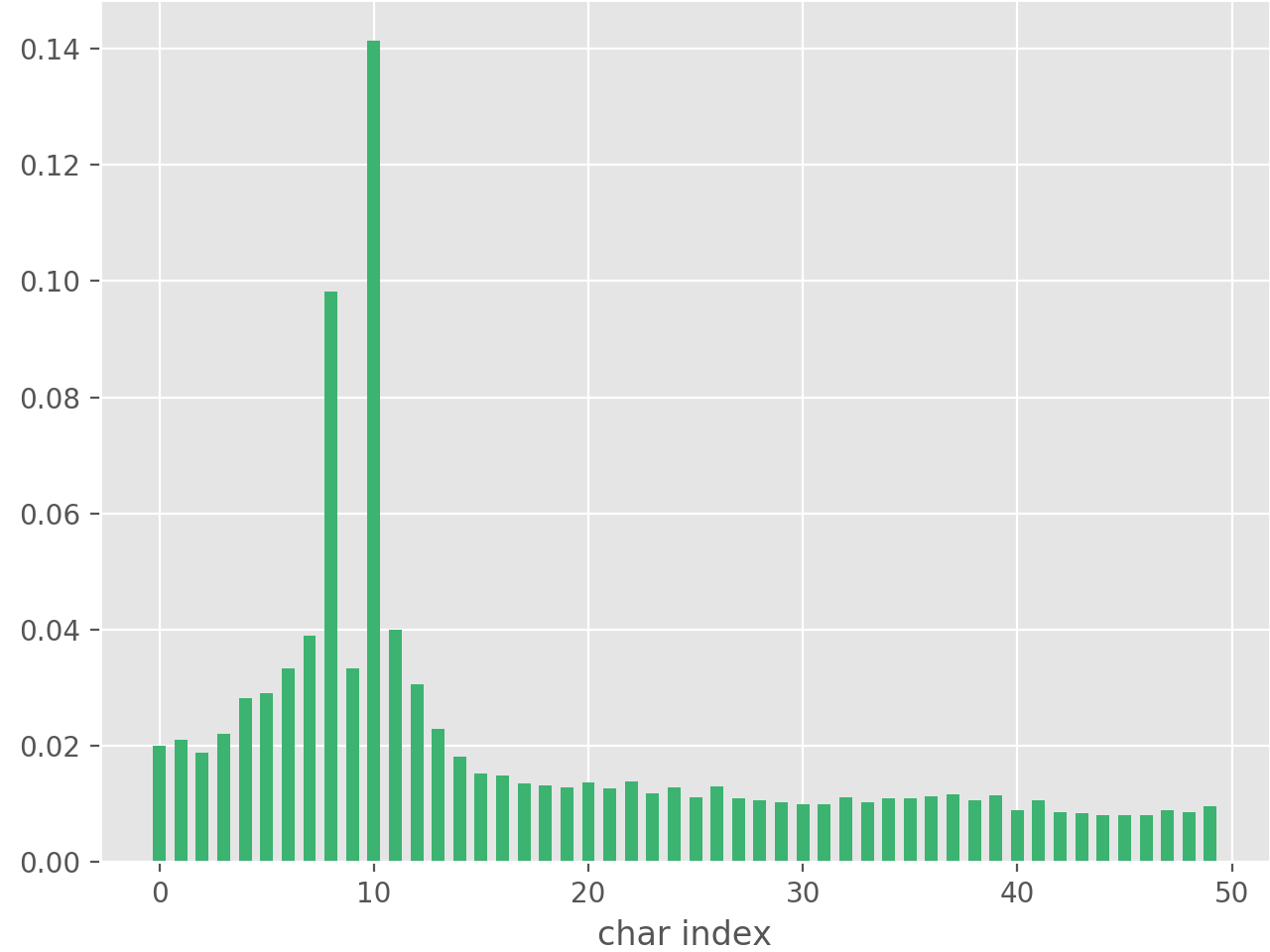} &   \includegraphics[width=0.4\textwidth]{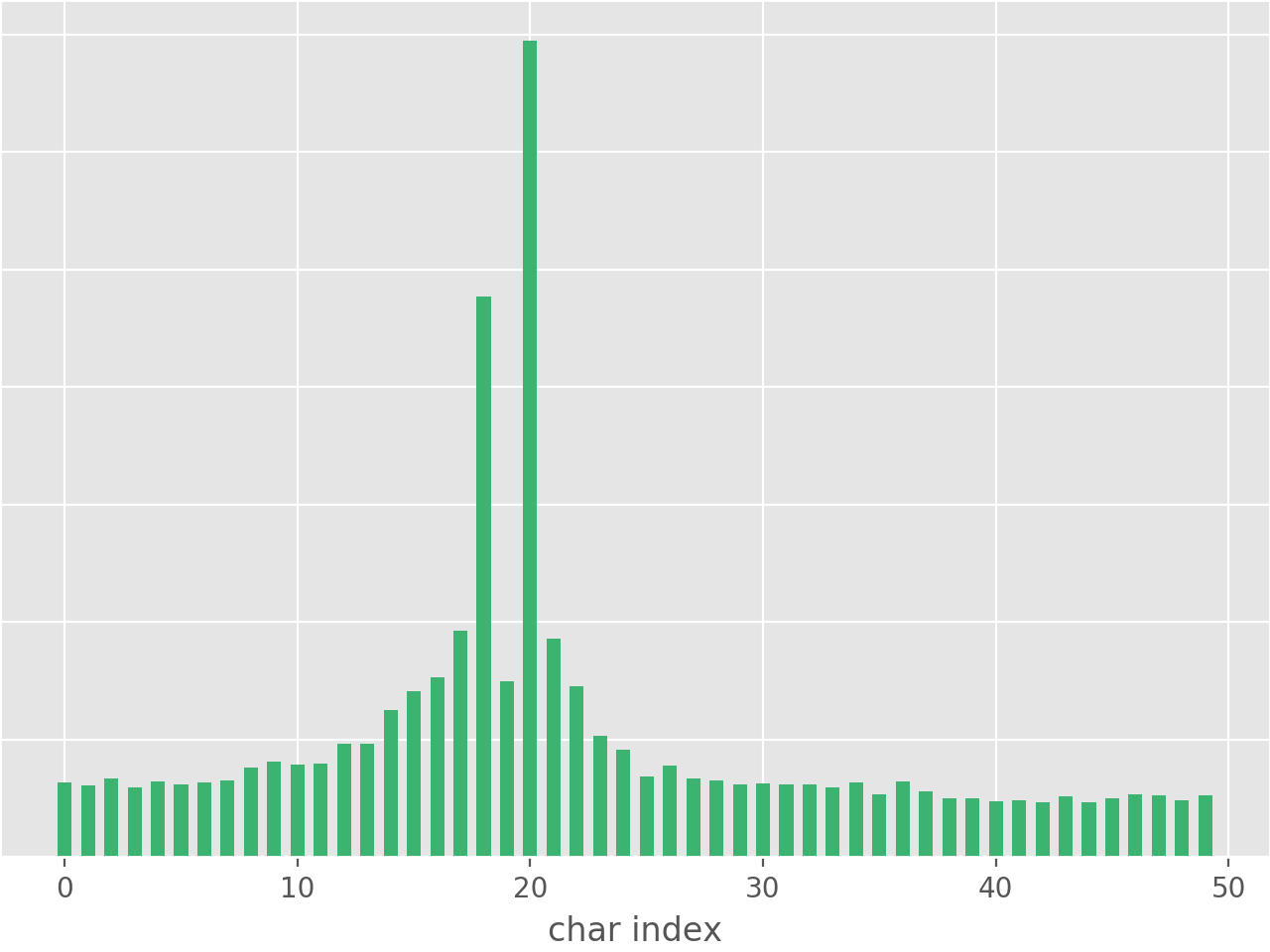} \\	
			(a) Current char index 10 & (b) Current char index 20  
			\\[6pt]
			\includegraphics[width=0.4\textwidth,,height=0.3\textwidth]{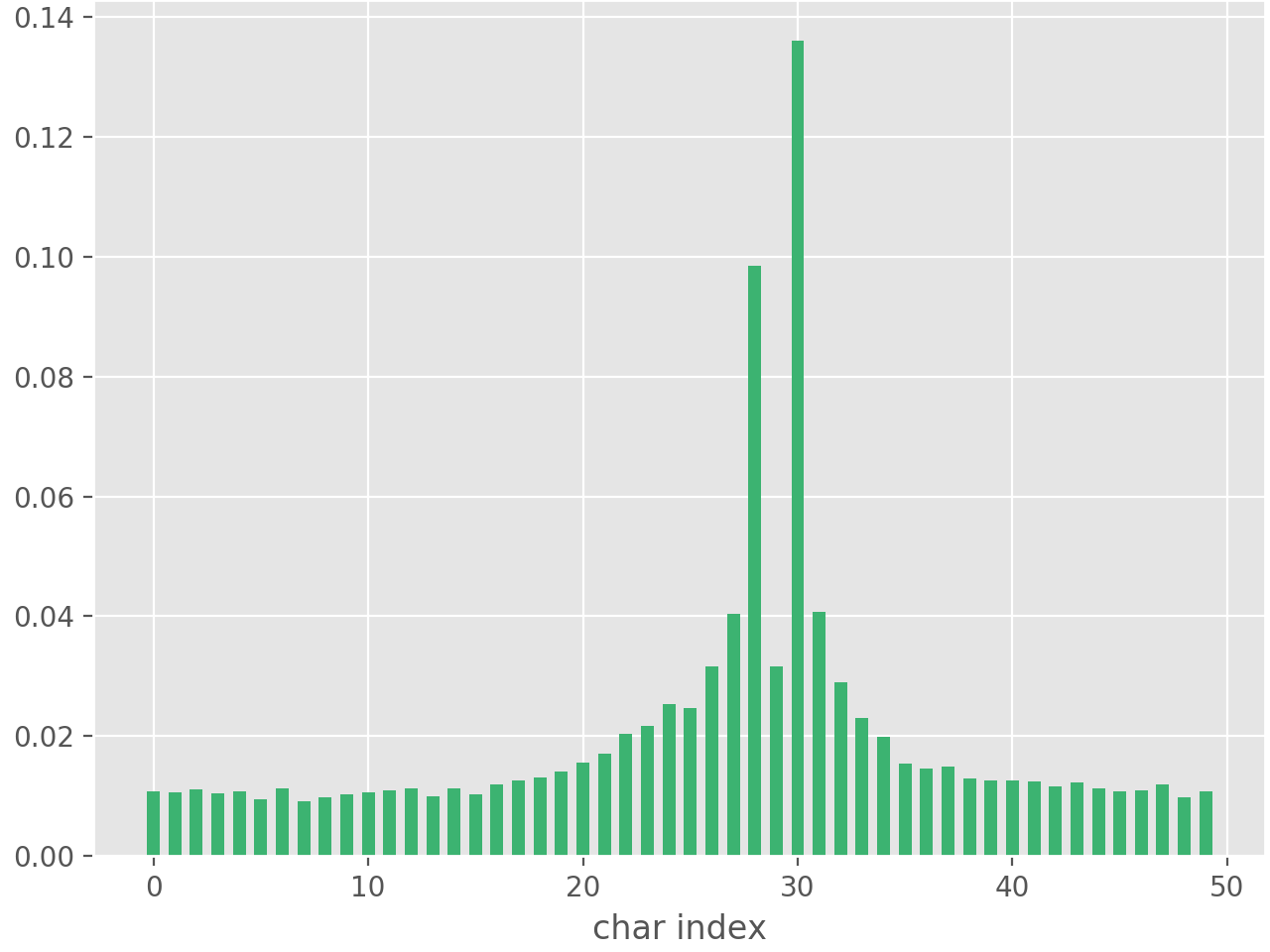} & \includegraphics[width=0.4\textwidth,height=0.3\textwidth]{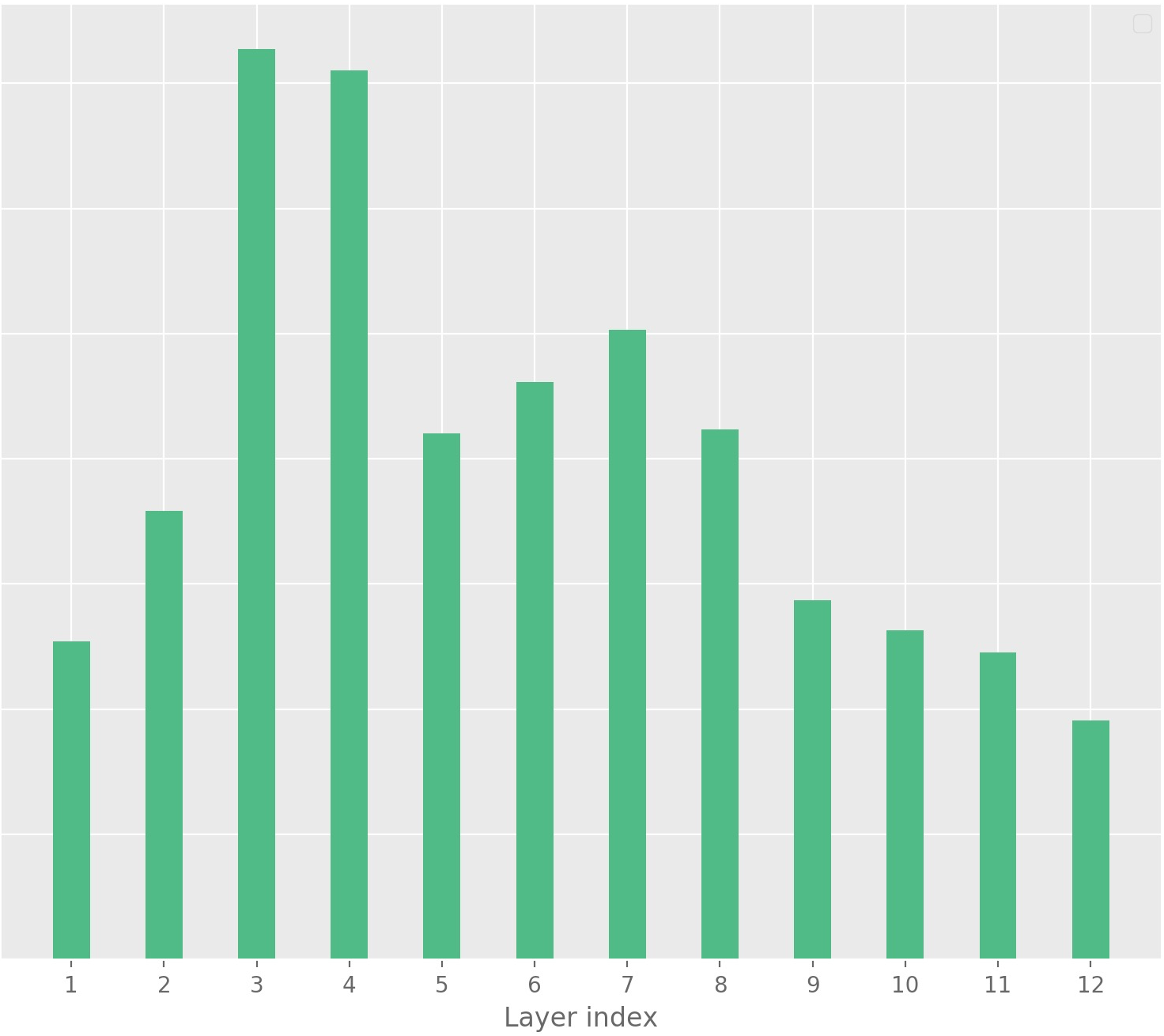} \\
			(c) Current char index 30 & (d) Layer attention score  \\[6pt]
		\end{tabular}
		\vspace*{-4mm}
		\caption{Distribution of self-attention score at certain char index (10, 20, 30) and layer attention score.}
		\label{self-attention}
	\end{figure*}
	
\subsection{Visualization}
	\label{sec:vis}
	\subsubsection{Self Attention Score Visualization}
	In the self-attention layers of the transformer, the attention score of each character is calculated with the rest characters. By visualizing the attention score, we can intuitively see what each character pay attention to. Specifically,  we choose the sentences with a length larger than 50 in all the ten datasets, feed these sentences into the model and average the attention scores at each index. The attention score is shown in Figure~\ref{self-attention}(a)(b)(c). We can notice that characters around the current character gain larger weights than those far away. The result indicates that word segmentation depends more on phrase-level information and long term dependencies are relatively unimportant. It intuitively proves that it is not necessary to keep long term memory of the sequence for CWS.
	
\subsubsection{Layer Attention Score Visualization}
\label{sec:layer_analysis}

BERT has achieved great success in many NLU tasks by pre-training a stack of 12 transformer layers to learn abundant knowledge. Intuitively, top layers capture high-level semantic features while bottom layers learn low-level features like grammar. As for chinese word segmentation task, high-level semantic features may have a small impact so that we make further investigation to find the minimal number of transformer layers. We freeze weights of each layer in the pre-trained BERT and conduct layer attention fine-tuning on word segmentation datasets. As shown in Figure~\ref{self-attention}(d), the attention score gradually decrease in top layers from 7 to 12, and \textbf{the third} layer gains the highest attention score. The results show that the model with three layers contains most information for word segmentation.  
Recent BERTology~\cite{jawahar2019does,liu2019linguistic,tenney2019bert,hewitt2019structural,goldberg2019assessing} analysis aimed to understand the inner working of BERT from the perspective of linguistics. These works came into a similar conclusion that the basic syntactic information of chunking appears in lower or middle layers, which is consistent with our analysis. The higher layers of BERT are the most semantics or tasks-specific. These analyses also reflect that Chinese word segmentation is a task of more syntactic but less partial semantic knowledge in linguistics.

	\section{Related Work}
	
	Recently, multi-criteria learning of neural CWS has drawn great attention of scholars. \cite{qiu2013joint}  adopted the stack-based model to take advantage of annotated data from multiple sources. \cite{chao2015exploiting} utilize multiple corpora using coupled sequence labeling model to learn and infer two heterogeneous annotations directly. ~\cite{gong2018switch}  proposed Switch-LSTM to improve the performance of every single criterion by exploiting the underlying shared sub-criteria across multiple heterogeneous criteria. ~\cite{chen2017adversarial} have proposed a multi-criteria learning framework for CWS. They proposed three shared-private models to integrate multiple segmentation criteria. An adversarial strategy is used to force the shared layer to learn criteria-invariant features. All these works utilize heterogeneous annotation data and show that they can indeed help improve each other.
	
	Model compression and acceleration in deep networks are vital in practice, which makes it possible to deploy deep models on mobile, embedded, and IoT devices. Techniques like parameter pruning, low-rank factorization, quantization and knowledge distillation had been widely used in visual tasks \cite{hinton2015distilling,ba2014deep,gupta2015deep,gong2014compressing}. However, model compression and acceleration are rarely investigated in NLP tasks, especially neural CWS task.  To the best of our knowledge, we are the first to compress the neural CWS model to accelerate the segmentation speed, by three model acceleration techniques, knowledge distillation, quantization and compiler optimization. We emphasize the segmentation speed is very important in industrial application, e.g., search engine.

	
	\section{Conclusion}
	In this paper, we propose an effective Chinese Word Segmentation method that employs BERT and adds a domain projection layer on the top with multi-criteria learning. They both serve to capture heterogeneous segmentation criteria and common underlying knowledge. And we visualize the attention score to illustrate linguistic within CWS. To be practicability, acceleration techniques are applied to improve the word segmentation speed.  It consists of knowledge distillation, quantization and compiler optimization.  Experiments show that our proposed model achieves higher performance on the word segmentation accuracy and faster prediction speed than the state-of-the-art methods.

\bibliographystyle{coling}
\bibliography{coling2020}

\end{document}